%% file: root.tex
\documentclass[letterpaper, 10 pt, conference]{ieeeconf}  

\IEEEoverridecommandlockouts                              

\usepackage{amsthm}

\usepackage{amsmath,amssymb,amsfonts}
\usepackage{bbm}
\usepackage{xcolor}
\usepackage{graphicx}
\usepackage[utf8]{inputenc}
\usepackage{booktabs,floatrow}
\usepackage[inkscapelatex=false]{svg}
\usepackage{caption}
\usepackage{subcaption}
\usepackage{url}
\usepackage{hyperref}
\usepackage[capitalise]{cleveref}
\usepackage{siunitx}
\usepackage[font=small]{caption} 
\usepackage{empheq}
\usepackage[most]{tcolorbox}

\usepackage[left=16.9mm,right=16.9mm,top=20.1mm,bottom=15.2mm]{geometry}

\usepackage[T1]{fontenc}

\renewcommand{\baselinestretch}{.9697} 

\newtcbox{\mymath}[1][]{%
    nobeforeafter, math upper, tcbox raise base,
    enhanced, colframe=black,
    colback=blue!5!white, boxrule=0.5pt,
    #1}

\newtheorem{definition}{Definition}
\newtheorem{assumption}{Assumption}

\title{\LARGE \bf
How Generalizable Is My Behavior Cloning Policy? A Statistical Approach to Trustworthy Performance Evaluation
}

\author{Joseph A. Vincent$^{1*}$, Haruki Nishimura$^{2}$, Masha Itkina$^{2}$, Paarth Shah$^{2}$, Mac Schwager$^{1}$, Thomas Kollar$^{2}$
\thanks{$^{*}$Work primarily done during internship at Toyota Research Institute.}
\thanks{The NASA University Leadership initiative (grant \#80NSSC20M0163) provided funds to assist the authors with their research, but this article solely reflects the opinions and conclusions of its authors and not any NASA entity. This research was partially supported by ONR grant N00014-23-1-2354. The first author was partially supported by the Dwight D. Eisenhower Transportation Fellowship.}%
\thanks{$^{1}$Department of Aeronautics and Astronautics, Stanford University, Stanford, CA 94305, USA, {\texttt\footnotesize \{josephav, schwager\}@stanford.edu}}%
\thanks{$^{2}$Toyota Research Institute, Los Altos, CA 94022, USA, {\texttt\footnotesize \{haruki.nishimura, masha.itkina, paarth.shah, thomas.kollar\}@tri.global}}%
}

\begin{document}

\maketitle
\thispagestyle{empty}
\pagestyle{empty}

\begin{abstract}
With the rise of stochastic generative models in robot policy learning, end-to-end visuomotor policies are increasingly successful at solving complex tasks by learning from human demonstrations.
Nevertheless, since real-world evaluation costs afford users only a small number of policy rollouts, it remains a challenge to accurately gauge the performance of such policies. This is exacerbated by distribution shifts causing unpredictable changes in performance during deployment.
To rigorously evaluate behavior cloning policies, we present a framework that provides a tight lower-bound on robot performance in an arbitrary environment, using a minimal number of experimental policy rollouts.
Notably, by applying the standard stochastic ordering to robot performance distributions, we provide a worst-case bound on the \textit{entire distribution} of performance (via bounds on the cumulative distribution function) for a given task. We build upon established statistical results to ensure that the bounds hold with a user-specified confidence level and tightness, and are constructed from as few policy rollouts as possible. In experiments we evaluate policies for visuomotor manipulation in both simulation and hardware. Specifically, we (i) empirically validate the guarantees of the bounds in simulated manipulation settings, (ii) find the degree to which a learned policy deployed on hardware generalizes to new real-world environments, and (iii) rigorously compare two policies tested in out-of-distribution settings. Our experimental data, code, and implementation of confidence bounds are open-source. \footnote{Project Page: \href{https://tri-ml.github.io/stochastic_verification/}{\text{https://tri-ml.github.io/stochastic\_verification}} \\ Paper Repository: \href{https://github.com/TRI-ML/stochastic_verification}{\text{https://github.com/TRI-ML/stochastic\_verification}} \\ Binomial Confidence Intervals: \href{https://github.com/TRI-ML/binomial_cis}{\text{https://github.com/TRI-ML/binomial\_cis}}}
\end{abstract}

\input{introduction}

\input{related_work}
\input{distributional_bounds}

\input{prob_form_binary}

\input{prob_form_continuous}

\input{experiments}

\section{Conclusion}
\label{sec:conclusion}

\textcolor{black}{The primary contribution of this paper is utilizing uncommon but optimal statistical bounds to efficiently measure policies' OOD generalization to unseen environments from few samples.
For binary and continuous performance metrics, we place confidence bounds on the entire distribution (i.e. CDF) of a policy's performance. 
To this end, a secondary contribution of this paper is the computation of MES for the UMA and Clopper-Pearson lower confidence bounds, which has applicability beyond robotics.
The final contribution of our paper is an open-source implementation of the UMA lower confidence bound and associated MES computation, both of which were absent from existing statistical software.}
Finally, in three experiments we show both the validity of our approach as well as how the approach can be used to make judgements about policy generalization in OOD settings. \textcolor{black}{Although we focus our discussion and experiments on BC policies, our approach can be applied to any black-box policy.} 

\textcolor{black}{The bounds presented in this paper are valid for the environment the samples were drawn from. One can extend the analysis in~\cite{vincent2023guarantees} to understand the sensitivity of the bounds' confidence level to distribution shifts, or to construct bounds which are robust to prescribed levels of distribution shift.} 
In our paper, the number of policy rollouts is determined by a budget or by constraints on confidence and tightness. However, our ideas can be extended to use confidence sequences where the number of rollouts is not determined beforehand~\cite{howard2022sequential}. It may save on testing costs to stop when enough evidence has been collected to conclude that the policy surpasses some performance threshold. Another direction is to evaluate performance during a single policy rollout, rather than multiple i.i.d. rollouts. As a final remark, when performing statistical evaluations, researchers should be forthright in their data collection, assumptions/modeling, and interpretation of results. 


\section*{Acknowledgment}
We thank Cheng Chi for guidance on diffusion policies.

\bibliographystyle{./IEEEtran} 
\bibliography{./IEEEabrv,./IEEEexample}


\addtolength{\textheight}{-12cm}   


\end{document}

%% file: introduction.tex
\section{Introduction} \label{sec:intro}



In this paper we focus on evaluating robot policies obtained through the \textit{behavior cloning}~(BC) framework of robot learning. BC policies, such as diffusion policies~\cite{chi2023diffusion}, have recently advanced the state-of-the-art (SOTA) in visuomotor policy synthesis, particularly in manipulation~\cite{chi2023diffusion,jang2022bc,fu2024mobile,zhao2023learning,florence2022implicit}. BC is attractive because it avoids the challenges of the sim-to-real gap, which impede the transfer of policies from reinforcement learning and other techniques to real-world robots~\cite{zhao2020sim}. In BC, humans give demonstrations, often directly on the robot hardware through teleoperation. 
Then, a policy is learned based on these demonstrations. This procedure removes the need for a simulation model, avoiding the sim-to-real gap. 

\begin{figure}
    \centering
    \includegraphics[width=0.9\linewidth]{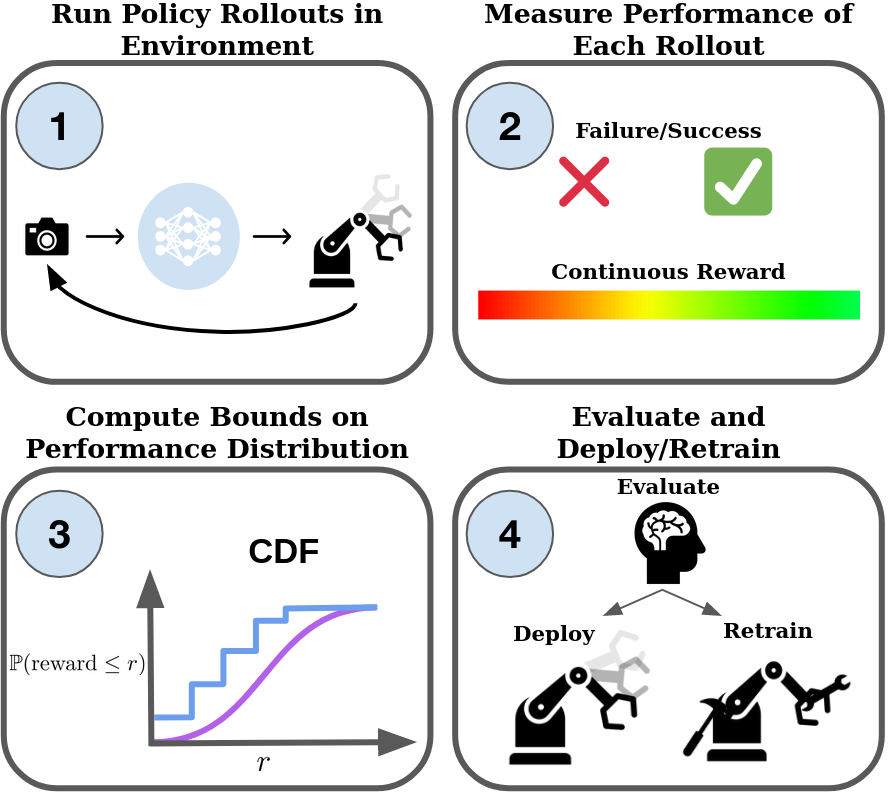}
    \caption{Our approach to evaluating BC policies. First, policy rollouts are collected in the environment of interest. Second, each rollout results in either a binary or continuous performance measurement. Third, statistical tools are used to compute an upper confidence bound on the CDF of performance. 
    Finally, the user interprets the confidence bound and chooses to deploy or retrain the policy.}
    \label{fig:first_fig}
\end{figure}

However, without an accurate simulation model, evaluation of BC policies relies on real-world tests. In robotics research it is common to evaluate these policies using fewer than $50$ policy rollouts, recording the empirical success rate or average reward (e.g.,~\cite{chi2023diffusion,jang2022bc, levine2016end}). With such small sample sizes, it can be difficult to interpret the significance of the recorded results. In addition, measuring average performance can be insufficient for applications with safety and reliability requirements.


To address this need, we propose statistical bounds to rigorously evaluate the performance of a BC policy. We quantify performance through a user-specified metric, either a binary success/failure or a continuous reward. Although many BC policies are trained and deployed in hardware, these metrics can also be given for simulated robotics settings. While specifying a suitable reward for policy training can be challenging in practice, we emphasize that we do not use the performance metric for training, only for evaluation. 
Simple performance metrics (e.g., task success/failure, distance to a goal location) are sufficient for our purposes. 

\begin{figure}
    \centering
    \includegraphics[width=0.99\linewidth]{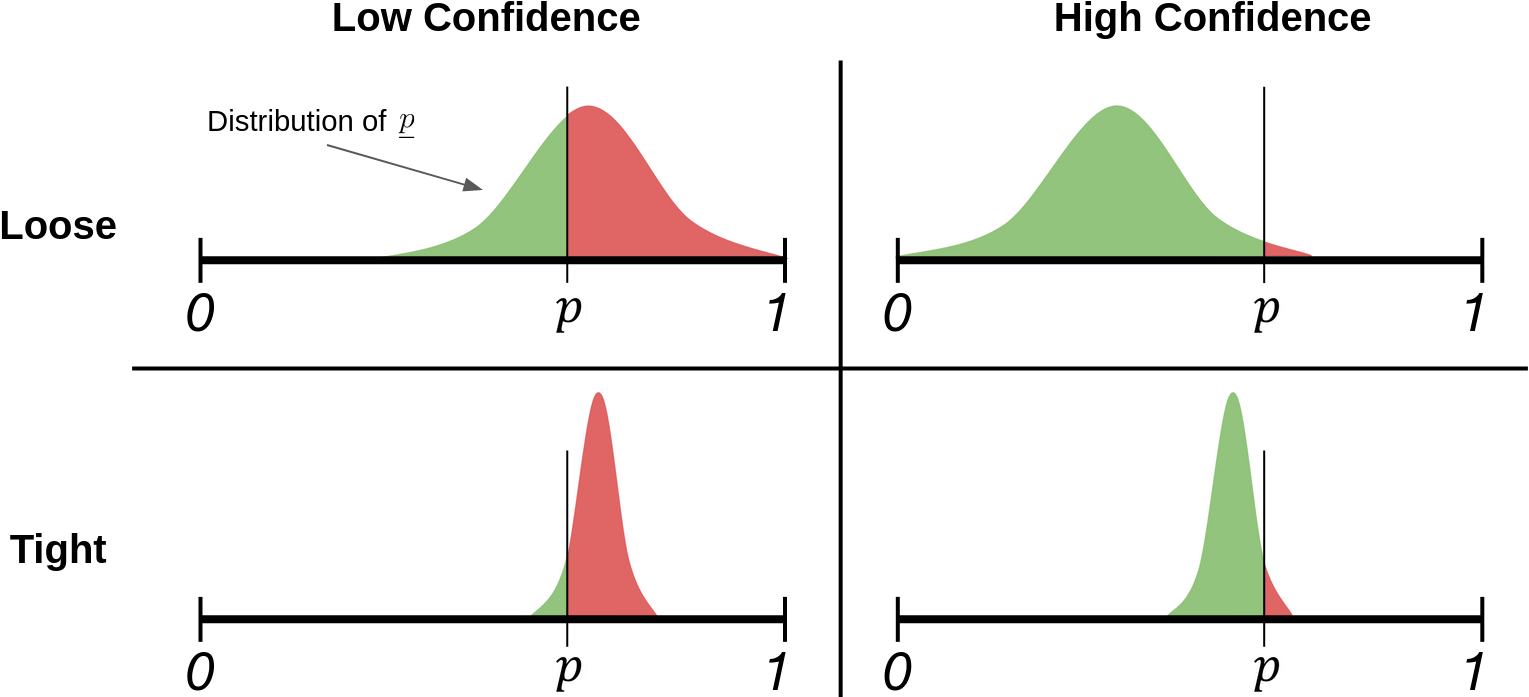}
    \caption{Hypothetical distributions of a lower confidence bound $\underline{p}$ for an unknown probability of success $p$. High confidence levels give better chances that the $\underline{p}$ we obtain is lower than $p$ (green shaded region), and tighter bounds give better chances that $\underline{p}$ is close to $p$.}
    \label{fig:conf_tightness}
\end{figure}

Our proposed method is to compute worst-case bounds on the performance of a policy based on the results of a small number of policy rollouts. \textcolor{black}{In \cref{sec:dist_bounds}, we define an ordering of distributions of policy performance. A worst-case bound on the performance of a policy is the least-preferable distribution according to this ordering that is consistent with the observed data (up to some confidence level).} Worst-case bounds are useful for determining whether a policy exceeds some performance threshold rather than determining it does not. Our approach can specify the fewest number of policy rollouts required to obtain a user-specified \emph{confidence} and \emph{tightness} for a bound on the performance metric. \emph{Confidence} is the probability (fraction of random outcomes) for which the bound holds.\footnote{Sometimes also referred to as the \textit{coverage} of the bound.} \textit{Tightness} quantifies how close the bound tends to be to the true (but unknown) quantity of interest. A bound which is tighter provides a better estimate of the unknown quantity while a bound which is higher confidence provides a better chance that the estimate is conservative (rather than optimistic). In \cref{fig:conf_tightness} we give a visual interpretation of how confidence and tightness relate to the distribution of a lower confidence bound for an unknown probability of success. Achieving higher confidence and tighter bounds on policy performance comes at the cost of more policy rollouts. 

The bounds we use require no knowledge of the underlying probability distributions in the policy, training data, robot hardware, or environment physics. The confidence and tightness of these bounds do not depend on the dimensionality of the robot state or observation. We only require performance measurements to be i.i.d. which we further discuss in \cref{sec:ass}.

In this paper we demonstrate how to construct worst-case bounds on performance when performance is measured by (i) a binary success/failure metric, and (ii) a continuous metric which measures the reward accumulated along a trajectory. For a binary metric, we seek a lower bound on the unknown probability of task success.  We revisit a rarely-used bound from the statistics literature, a tighter version of the more common Clopper-Pearson bound~\cite{clopper_pearson}. Our bound exactly achieves a user-specified confidence level and is as tight as possible.
This is enabled by a novel computational method based on mixed monotonic optimization~\cite{matthiesen2020}.
For a continuous metric, the true distribution of reward is entirely unknown. We propose to use bounds for the cumulative distribution function (CDF) of reward. The CDF fully characterizes the reward distribution, allowing flexibility for the user to analyze performance, e.g., in the tails, at any quantile, or in the mean.  We use a tighter version of the Dvoretzky–Kiefer–Wolfowitz (DKW) bound \cite{dvoretzky1956,massart1990} for this purpose.  
In both the binary and continuous cases, the user specifies the confidence level and the tightness of the bound. Since real-world evaluation is costly, the bounds we employ meet these specifications using the minimum number of policy rollouts. This is in contrast to the more common Clopper-Pearson and DKW bounds.



Finally, in our experiments we are primarily interested in applying our framework to investigate the generalization of BC policies to out-of-distribution (OOD) settings. \textcolor{black}{We accomplish this by directly testing in the OOD settings.} Because human intuition is often unreliable for predicting which OOD settings will be harmful to the policy, we believe this is an application where statistical evaluation methods shine, giving roboticists well-defined confidence levels for the results of their experiments. We demonstrate our bounds in simulation and real-world manipulation settings. We compare our bounds to more common bounds from the literature, showing improved tightness, particularly for few policy rollouts. We also show how our bounds can be used to compare the performance of two policies. Finally, in hardware, we bound the performance of a SOTA visuomotor diffusion policy~\cite{chi2023diffusion} in new environments. An early version of these results appeared in a workshop~\cite{vincent2023full}. Here we substantially expand the analysis and simulation results, add hardware experiments, and add a policy comparison example. Our contributions are as follows:
\begin{itemize}
    \item A method for bounding the performance of a BC robot policy in an arbitrary environment with minimal policy rollouts, by putting in practice the concept of worst-case bounds on the entire distribution of performance.
    \item A novel method for computing tightness for binomial confidence intervals, specifically \textit{maximum expected shortage} (the worst-case value for the average degree of underestimation of the true success rate).
    \item An open-source implementation for computing the binomial bounds described in the paper, which optimally trade-off between confidence, tightness, and sample size: \href{https://github.com/TRI-ML/binomial_cis}{\ \ \ \ \text{https://github.com/TRI-ML/binomial\_cis}}.
\end{itemize}

%% file: related_work.tex
\section{Related Work} \label{sec:related}
In this section, we discuss related literature on rigorously evaluating learned robot policies. We first discuss approaches from formal methods, then discuss statistical approaches.

\subsection{Formal Evaluation of Learned Policies}
Formal methods applied to robots with learned components can be used to evaluate open-loop policy requirements (such as state-dependent action constraints~\cite{katz2017reluplex}) and closed-loop system requirements (such as reach-avoid constraints~\cite{tran2020nnv, vincent2022reachable, katz2022verification, rober2023backward}). These methods typically have strict limitations on the structure and complexity of the policy and environment model. In addition, these methods are most commonly employed in settings with deterministic environments or with bounded (set-based) uncertainties. In settings where these requirements are met, formal methods can evaluate rich closed-loop behaviors by finding certificates such as invariant sets~\cite{vincent2022reachable, katz2022verification}, Lyapunov functions~\cite{richards2018lyapunov, dai2021lyapunov}, and contraction metrics~\cite{singh2021learning, sun2021learning}. However, in this paper we do not assume access to an environment model or any particular policy structure, precluding the use of formal evaluation methods. When systems become too complex for formal evaluation, black-box validation methods are used~\cite{corso2021survey}. Statistical testing, discussed next, is one flavor of black-box validation that provides probabilistic guarantees.


\subsection{Statistical Evaluation of Learned Policies}
When constructing statistical bounds, there are trade-offs between confidence, tightness, and sample size. Related work tends to focus on the trade-off between confidence and number of samples, neglecting the important role that tightness plays in interpreting the bound. Placing statistical bounds on the performance of learned policies from few rollouts is explored in~\cite{caltech_verification, caltech_policy_synth, risk_verification, vincent2023guarantees}. However, these works only bound particular scalar quantities such as the expected value, value-at-risk, and conditional value-at-risk of the performance distribution. We place bounds on the CDF that consider all aspects of the performance distribution. For example, policies with appealing expected performance may have unacceptable long tails of poor performance. Further, these papers do not quantify bound tightness and their bounds can require more policy rollouts than necessary to achieve the specified confidence.

Conformal prediction has been used for uncertainty quantification of robot perception components~\cite{luo2022sample, lindemann2023safe, dixit2023adaptive, ren2024explore} and policies~\cite{ren2023robots, wu2023uncertainty}. This research has enabled policies to attain a specified probability of task success, such as avoiding collisions or making language-based plans. However, like the other approaches in this section, (i) by bounding a quantile of the score distribution, conformal prediction bounds also ignore tails of distributions, (ii) tightness of bounds from conformal prediction is often not quantified, and (iii) conformal prediction may require more samples than necessary to achieve a desired confidence~\cite{papadopoulos2011regression, angelopoulos2021gentle}. In this paper we bound the entire distribution of performance, quantify bound tightness, and ensure that the number of samples used is minimal.

Confidence bounds for unknown success probabilities and CDFs are well studied with the Clopper-Pearson bound being common for success probabilities~\cite{clopper_pearson} and with the DKW inequality being common for CDFs~\cite{dvoretzky1956,massart1990}. However, these bounds are loose and may require more policy rollouts than necessary. In \cref{sec:binary} and \cref{sec:continuous}, we discuss less common yet well-established confidence bounds which optimally trade-off between confidence, tightness, and number of samples. Lastly, in \cref{sec:comparison_ex} we compare the success rates of two policies using the bounds in this paper. Common approaches which test for differences in success rates include Fisher's and Barnard's tests~\cite{two_sample}. \textcolor{black}{These are the standard methods for performing A/B tests to distinguish success rates.} However, while such tests determine which policy performs better, they do not return confidence intervals for the individual success rates (as we do), but rather provide confidence intervals for the odds ratio. \textcolor{black}{The drawback is that the odds ratio lacks information about absolute performance, i.e., pairs of policies with low success rates and with high success rates can have the same odds ratio.} Our view is that it is not enough to know which policy performs better, but it is also important to know the performance of each policy individually.

%% file: distributional_bounds.tex
\section{Distributional Bounds} \label{sec:dist_bounds}
\subsection{Bounds}
We want to obtain worst-case bounds on the \textit{entire distribution} of performance (via bounds on the CDF), rather than conventional quantities such as expected value. This approach affords a rich understanding of performance, hedging against deploying a policy with some hidden drawbacks (e.g., high expected reward but a long tail of low reward).


Consider two CDFs $F_1(x) = \mathbb{P}(X_1\le x)$ and $F_2(x) = \mathbb{P}(X_2\le x)$, describing  the distribution of performance under two different policies. If $F_2(x) \ge F_1(x) \ \forall x$, then $F_2(x)$ has more probability mass on \emph{lower} performance values. Therefore $F_1(x)$ is preferable to $F_2(x)$, following the standard notion of stochastic ordering~\cite{lehmann_textbook}. If the inequality holds for some, but not all, values of $x$, we do not claim one distribution is preferable to the other. Therefore we have a partial ordering of performance based on CDFs. 

An upper confidence bound $\overline{F}(x)$ on a CDF with confidence level $1-\alpha$ satisfies
\begin{align}
    \mathbb{P}(F(x) \le \overline{F}(x) \ \forall x) \ge 1-\alpha. \label{eq:gen_bound}
\end{align}
$\overline{F}(x)$ is constructed based on the observed performance of some number of policy rollouts. Since the outcome of policy rollouts is random, $\overline{F}(x)$ is also random, leading to the bound holding probabilistically as in \cref{eq:gen_bound}.
In light of the above discussion on partial ordering of CDFs, $\overline{F}(x)$ is a \emph{worst-case} bound on the distribution of performance with confidence $1-\alpha$.  In the case of a binary performance metric, an upper bound on the CDF is simply a lower bound on the success rate.

\subsection{\textcolor{black}{Assumptions}} \label{sec:ass}
\textcolor{black}{
The methods we use in \cref{sec:binary,sec:continuous} to construct $\overline{F}(x)$ rely on i.i.d. performance measurements.
\begin{assumption} \label{assumption}
    The performance measurements from the policy rollouts are i.i.d. random variables.
\end{assumption}
That the performance measurements are random variables is ensured by a stochastic policy and/or stochastic environment. To be independent, the outcome of one policy rollout must have no influence on the outcome of another. To be identically distributed, there must be no distribution shift between rollouts. For binary metrics, \cref{assumption} ensures the performance will follow a Bernoulli distribution with unknown probability of success. For continuous metrics, the assumption allows for the performance to follow any distribution (including discrete and mixed continuous-discrete distributions).}

\textcolor{black}{To avoid violating \cref{assumption}, a testing plan should be developed \textit{before} collecting any policy rollouts, documenting the sampling rule for initial conditions and the number of policy rollouts. Practices which violate \cref{assumption} include
\begin{itemize}
    \item running policy rollouts until enough favorable outcomes are observed (in which case confidence sequences would be more appropriate~\cite{howard2022sequential}),
    \item running policy rollouts in a time-varying environment (e.g., lighting changes between rollouts due to sunset).
\end{itemize}
For further discussion on misuse of statistical evaluation see \cite{greenland2016statistical}. Next, we give worst-case bounds for both the binary and continuous performance metrics.}

%% file: prob_form_binary.tex
\section{Confidence Bounds - Binary Metric} \label{sec:binary}
We are interested in finding lower confidence bounds on the probability of success that require a minimal number of policy rollouts while achieving user-specified levels of confidence and tightness. To determine such bounds we use the approach in \cite{lehmann_textbook} to achieve a desired confidence level with minimal rollouts. We then introduce a computational method to achieve a desired tightness with minimal rollouts.

To find lower bounds on a policy's success rate in a new environment, we treat the result of each policy rollout as a Bernoulli random variable $X$, where $X=1$ denotes a success and $X=0$ denotes a failure. Then, given the $n$ i.i.d. success/failure measurements,  we construct a lower confidence bound $\underline{p}$. The number of observed successes follows a binomial distribution. Before describing the bound, we first introduce optimality criteria for lower confidence bounds.

\subsection{Optimality Criteria}
The standard notion of optimality for lower confidence bounds is that of being \textit{uniformly most accurate} (UMA).
\begin{definition}[Uniformly Most Accurate, Eq. 3.22 of \cite{lehmann_textbook}]
    For test statistic $T$, a lower confidence bound $\underline{\theta}(T)$ satisfying
    \begin{gather}
        \mathbb{P}_\theta [\underline{\theta}(T) \le \theta] \ge 1-\alpha \quad \forall \theta
    \end{gather}
    and, amongst all possible lower bounds $\underline{\theta}^\prime(T)$, minimizes
    \begin{gather}
        \mathbb{P}_\theta [\underline{\theta}^\prime(T) \le \theta_0] \quad \forall \theta_0 < \theta
    \end{gather}
    is a UMA lower confidence bound at confidence level $1-\alpha$.
\end{definition}
Intuitively, a UMA lower confidence bound underestimates the unknown parameter $\theta$ by as little as possible, no matter what the actual value of $\theta$ is.  \textit{Shortage} (i.e. excess width) is used to quantify the amount of underestimation,
\begin{align}
    \text{shortage} = \max\{\theta - \underline{\theta}, 0\}.
\end{align}
Using the Ghosh-Pratt identity~\cite{ghosh1961, pratt1961}, 
   \begin{gather}
    \textup{ES}(\theta) = \mathbb{E}_\theta[\textup{shortage}] = \int_{\theta_0 < \theta} \mathbb{P}_\theta[\underline{\theta} \le \theta_0] d\theta_0. \label{eq: es_theta}
\end{gather} 
Now, since a UMA lower confidence bound minimizes the integrand of \cref{eq: es_theta}, a UMA lower confidence bound also minimizes expected shortage for all values of $\theta$. 
Although expected shortage is a useful quantity, it depends on the unknown value of $\theta$. To avoid assumptions on $\theta$, we measure tightness according to \textit{maximum expected shortage} (\textit{MES}),
\begin{align}
    \textup{MES} = \textcolor{black}{\underset{\theta}{\text{maximize}} \ \ \textup{ES}(\theta)}. \label{eq: MES}
\end{align}
MES is the worst-case expected shortage over all possible parameter values $\theta$. The notion of MES is not new~\cite{edwardes1998}, but it is not commonly used since the maximization in \cref{eq: MES} can be challenging when expected shortage is not concave in $\theta$. Next, we describe a UMA lower confidence bound for an unknown success probability and introduce the first tractable method for computing the associated MES.


\subsection{Optimal Binomial Bounds}
In this paper we let $\textup{bin}(k,n,p)$ and $\textup{Bin}(k,n,p)$ denote the binomial probability mass function (PMF) and CDF, with $k$ successes, $n$ trials, and success probability $p$. The floor function $\lfloor x \rfloor$ rounds the argument $x$ down to the nearest integer. Now, we define a test statistic $T = U + \sum_{i=1}^n X_i$ where $U \sim \mathcal{U}[0,1]$ and $X_i$ are the observed successes and failures. Then $T$ is random with probability density
\begin{align}
    f_p(t) = {n \choose \lfloor t \rfloor} p^{\lfloor t \rfloor} (1-p)^{n - \lfloor t \rfloor}.
\end{align}


Then, we can compute the CDF as follows,
\begin{subequations} \label{eq: cdf}
   \begin{gather}
    F_p(t) = \int_0^{t} {n \choose \lfloor x \rfloor} p^{\lfloor x \rfloor} (1-p)^{n - \lfloor x \rfloor} dx \\
    = \textup{Bin}(\lfloor t \rfloor - 1, n, p) + (t - \lfloor t \rfloor) \cdot \textup{bin}(\lfloor t \rfloor, n, p). 
\end{gather} 
\end{subequations}
$F_p(t)$ is continuous in both $t$ and $p$. 
Then, by Corollary 3.5.1 of~\cite{lehmann_textbook}, we have the UMA lower confidence bound,
\begin{subequations}
    \begin{empheq}[box=\mymath]{gather}
    \mathbb{P}[\underline{p} \le p] \ge 1-\alpha \\
    \underline{p}(t) = \begin{cases}
                        0 \quad &\text{if } t < 1-\alpha \\
                        1 \quad &\text{if } t > n + 1-\alpha \\
                        p^* \quad &\text{s.t. } F_{p^*}(t) = 1-\alpha \text{ otherwise.}
    \end{cases}
    \label{eq:p_lo}
\end{empheq}
\end{subequations}
\textcolor{black}{Note that since the test statistic is defined as a sum of uniform and binomial random variables, given the same number of observed successes, $t$ will be almost surely distinct, resulting in distinct values for $\underline{p}(t)$. Thus, $\underline{p}$ is known as a \textit{randomized confidence bound}, and this property is necessary to construct a UMA confidence bound for an unknown success rate.}

A bisection search can be used to solve \cref{eq:p_lo}, as $F_p(t)$ is decreasing in $p$. Taking limits $p \rightarrow 0^+$ and $ p \rightarrow 1^-$, one finds $F_p(t) = 1-\alpha$ has no solution when $t<1-\alpha$ and $t>n+1-\alpha$. This bound is well-established (see Example 3.5.2 of~\cite{lehmann_textbook}). The Clopper-Pearson bound is obtained by setting $U = 0$ in the test statistic.





Next, we compute expected shortage. A useful identity is
\begin{align}
    \mathbb{P}[\underline{p} \le p_0] = \mathbb{P}[t \le t^*(p_0)]
\end{align}
where $t^*(p_0)$ is the unique value that satisfies
\begin{gather}
    F_{p_0}(t^*) = 1-\alpha.
\end{gather}
Then, we can compute expected shortage as
\begin{subequations}
    \begin{gather}
        \textup{ES}(p) = \int_0^p \mathbb{P}_p[\underline{p} \le p_0] dp_0 = \int_0^p \mathbb{P}_p[t \le t^*(p_0)] dp_0 \\
        = \int_0^p F_p(t^*(p_0)) dp_0.
    \end{gather} \label{eq: es}
\end{subequations}
This integral can be evaluated with standard software.

Next, we find MES via global optimization. Specifically, consider the following reformulation of expected shortage,
\begin{gather}
    \textup{ES}(p_1, p_2) = \int_0^{p_1} F_{p_2}(t^*(p_0)) dp_0.
\end{gather}
In this form, expected shortage is increasing in $p_1$ and decreasing in $p_2$; this property is known as \textit{mixed monotonicity}. Mixed monotonic functions can be globally optimized~\cite{matthiesen2020}. This approach is the first tractable method that certifiably solves the optimization needed to compute MES and is included in our open-source implementation.\footnote{\href{https://github.com/TRI-ML/binomial_cis}{\ \ \ \ \text{https://github.com/TRI-ML/binomial\_cis}}}

\begin{figure}
    \centering
    \includegraphics[width=0.8\linewidth]{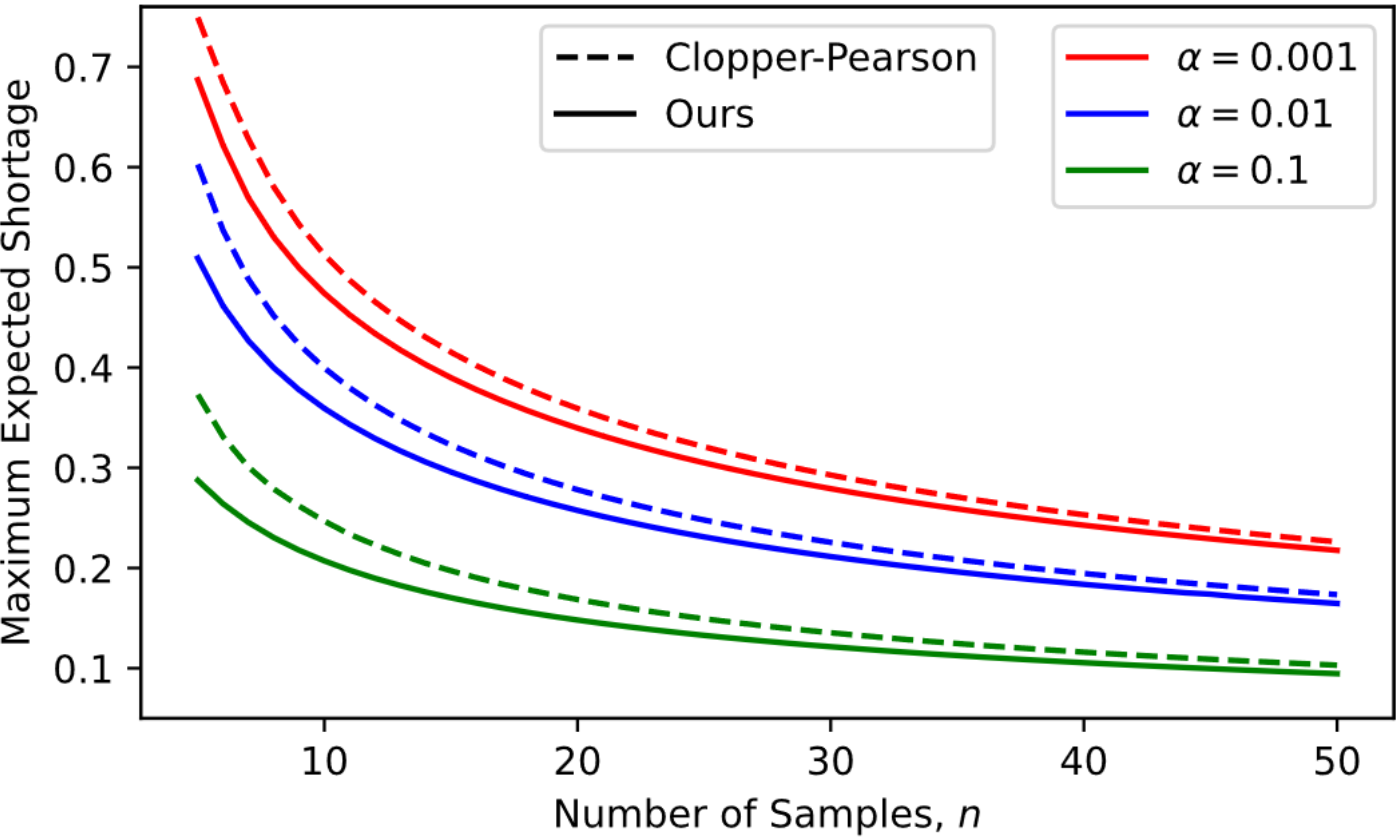}
    \caption{Variation of MES, $n$, and $\alpha$. Our method is always tighter than Clopper-Pearson, and appreciably so at small sample sizes.}
    \label{fig:MES_trade-off}
\end{figure}

Now, given two of the following: (i) confidence level ($1-\alpha$), (ii) tightness (MES), (iii) number of policy rollouts ($n$), we can determine the third. \textcolor{black}{For example, since MES is decreasing in $n$, to determine $n$ given a desired $\alpha$ and MES, one fixes $\alpha$ and computes the smallest $n$ (via a bisection search) which yields an MES below the desired value.} In \cref{fig:MES_trade-off} we show how these quantities vary with one another for our bound and the Clopper-Pearson bound. Since the bound we use is UMA, it always has lower MES. Also novel is the MES computation for Clopper-Pearson. We derive the appropriate form of \cref{eq: es} and compute MES using mixed monotonic programming.\footnote{\textcolor{black}{Further details on this derivation for the Clopper-Pearson bound are in the documentation of the linked \href{https://github.com/TRI-ML/binomial\_cis}{\text{binomial\_cis}} Github repository.}} Lastly, one can compute a lower confidence bound on the \textit{failure} probability using the same procedure, but with failure counts in the test statistic $T$ rather than success counts.

%% file: prob_form_continuous.tex
\section{Confidence Bounds - Continuous Metric} \label{sec:continuous}
When performance is measured with a continuous metric, the policy rollouts result in i.i.d. samples from some unknown distribution of reward. This distribution may be continuous, discrete, or mixed. In this section we describe how to obtain an upper confidence bound on the CDF of this distribution. As in \cref{sec:binary}, the CDF upper bound is constructed in such a way that given user-specified confidence level and tightness, the minimum number of policy rollouts is used to meet these specifications. The results used in this section are not novel (see~\cite{birnbaum1951}), but are uncommon in the statistics literature, especially as applied to robotics.

\begin{figure}
    \centering
    \includegraphics[width=0.8\linewidth]{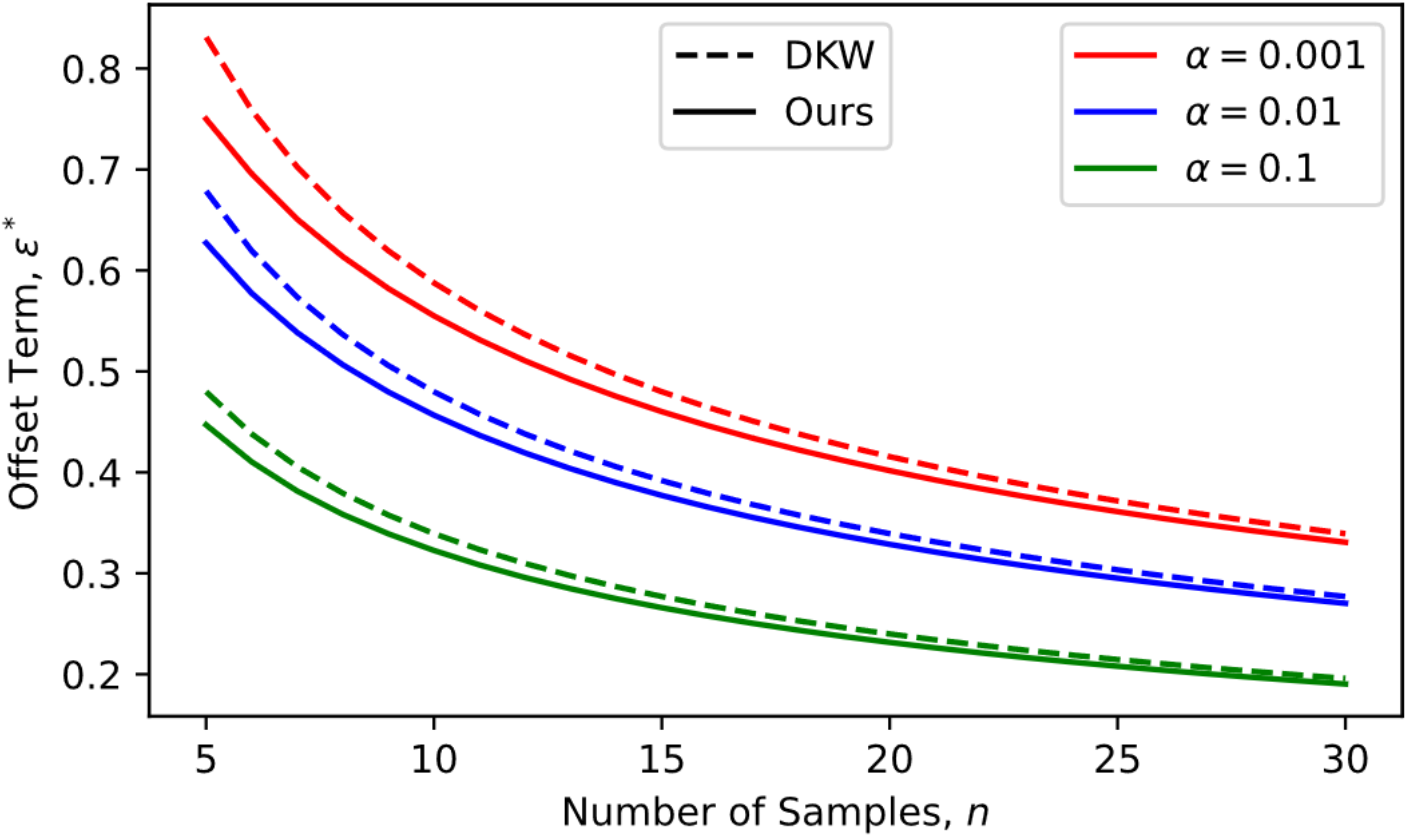}
    \caption{Variation of $\epsilon^*$, $n$, and $\alpha$. Our method is always tighter than the DKW bound, and appreciably so at small sample sizes.}
    \label{fig:cdf_comp}
\end{figure}

Consider an i.i.d. sample, $X_{1:n}$ from an unknown distribution with CDF $F(x)$. The empirical CDF is defined as
\begin{align}
    F_n(x) = \frac{1}{n} \sum_{i = 1}^n \mathbf{1}(X_i \le x),
\end{align}
where $\mathbf{1}(x)$ evaluates to $1$ if $x$ is true and $0$ otherwise. A one-sided Kolmogorov-Smirnov (KS) statistic~\cite{lehmann_textbook} is
\begin{align}
    D_n^- = \sup_x \  F(x) - F_n(x).
\end{align}
By~\cite{birnbaum1951}, if $F(x)$ continuous, then the distribution of $D_n^-$ is
\begin{subequations}
    \begin{gather}
    \mathbb{P}(D_n^- \le \epsilon) = 1 - \epsilon \sum_{k=0}^{\lfloor n(1-\epsilon) \rfloor} w_k \label{eq: KS_prob} \\
    w_k = {n \choose k} (1-\epsilon - \frac{k}{n})^{n-k} (\epsilon +\frac{k}{n})^{k-1}.
\end{gather} \label{eq: ks_dist}
\end{subequations}
Note that there is no dependence on the true $F(x)$. Moreover, if $F(x)$ is not continuous, then \cref{eq: KS_prob} holds with $\ge$ rather than equality. Next, note the equivalence
\begin{align}
    \sup_x \  F(x) - F_n(x) \le \epsilon \Leftrightarrow F(x) \le F_n(x) + \epsilon \ \forall x.
\end{align}
Then, given i.i.d. samples $X_{1:n}$ from some $F(x)$, we have the following upper confidence bound on the CDF~\cite{birnbaum1951},
\begin{subequations}
   \begin{empheq}[box=\mymath]{gather}
   \mathbb{P}(F(x) \le \overline{F}(x) \ \forall x) \ge 1-\alpha \label{eq:cdf_ub_a} \\
    \overline{F}(x) = F_n(x) + \epsilon^* \label{eq:cdf_ub} \\
    \text{with $\epsilon^*$ chosen s.t.\ } \mathbb{P}(D_n^- \le \epsilon^*) = 1-\alpha. \label{eq: eps}
    \end{empheq}
\end{subequations}
The tightness of this bound is measured by $\epsilon^*$, the offset from the empirical CDF. The $\epsilon^*$ chosen by \cref{eq: eps} is optimal, if made any smaller then \cref{eq:cdf_ub_a} would not hold. $\epsilon^*$ is easily computed via a bisection search. 
\textcolor{black}{Since $\epsilon^*$ is decreasing in $n$, to determine $n$ given a desired $\alpha$ and $\epsilon^*$, one fixes $\alpha$ and computes the smallest $n$ (via a bisection search) which yields an $\epsilon^*$ below the desired value.}
One can also use the DKW inequality~\cite{dvoretzky1956, massart1990} to obtain a similar bound, except with $\epsilon^*~=~\sqrt{-\ln{\alpha} / (2n)}$. However, the DKW bound is always less tight as shown in \cref{fig:cdf_comp}.

%% file: experiments.tex
\begin{figure}[t!]
    \centering
    \includegraphics[width=\linewidth]{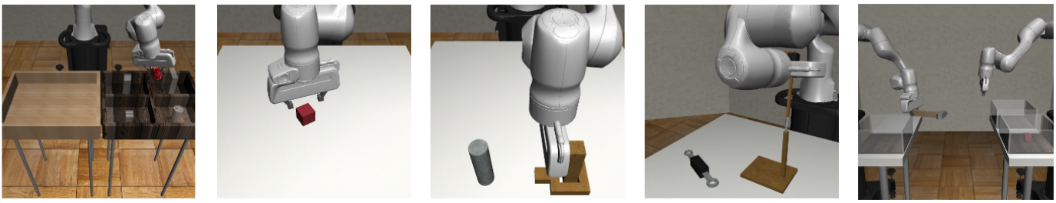}
    \caption{Simulation tasks: can, lift, square, tool hang, transport \cite{zhu2020robosuite}. }
    \label{fig:sim_snapshots}
\end{figure}
\section{Experiments} \label{sec: experiments}
We demonstrate our approach to evaluating BC policies in both simulation and real-world robotic manipulation settings. Performance of BC policies in OOD settings is hard to predict since it is difficult to characterize what aspects of the policy are invariant or sensitive to particular distribution shifts. 
This motivates the need for rigorous statistical evaluation. 
Thus, all of our experiments evaluate policies in OOD settings.
\textcolor{black}{In the absence of assumptions on the nature of the distribution shift, it is necessary to directly test the policies in the OOD settings.}

In simulation, we run many evaluations and show the empirical confidence and tightness of the bounds agree with the theory. In hardware, we (i) investigate the degree to which a learned policy generalizes to OOD settings, and (ii) compare the OOD generalization of two learned policies. Our approach requires i.i.d. performance scores; a sufficient condition to ensure this is for the initial observations to be i.i.d. from some (potentially unknown) distribution. In simulation, this is easily satisfied. In hardware, we took efforts to eliminate any time-varying conditions that may undermine this assumption. Because our bounds are strictly tighter than those of Clopper-Pearson and DKW, we report the results for our bounds only.

\subsection{Simulation Results} \label{sec:sim}
We consider five diffusion policies from~\cite{chi2023diffusion} for visuomotor manipulation that have shown recent SOTA performance, each trained for a separate task. \cref{fig:sim_snapshots} shows training environments. After training, we modified each environment as follows,
\begin{itemize}
    \item \textbf{Can}: Changed the color of the can to lime green, and the color of the floor to beige. 
    \item \textbf{Lift}: Changed the cube color to blue.
    \item \textbf{Square}: Changed the square color to tan.
    \item \textbf{Tool Hang}: Changed the wall color to floral white.
    \item \textbf{Transport}: Changed the color of the cube to lime green and the color of the lid handle to silver.
\end{itemize}

\begin{figure}[t!]
    \centering
    \includegraphics[width=0.98\linewidth]{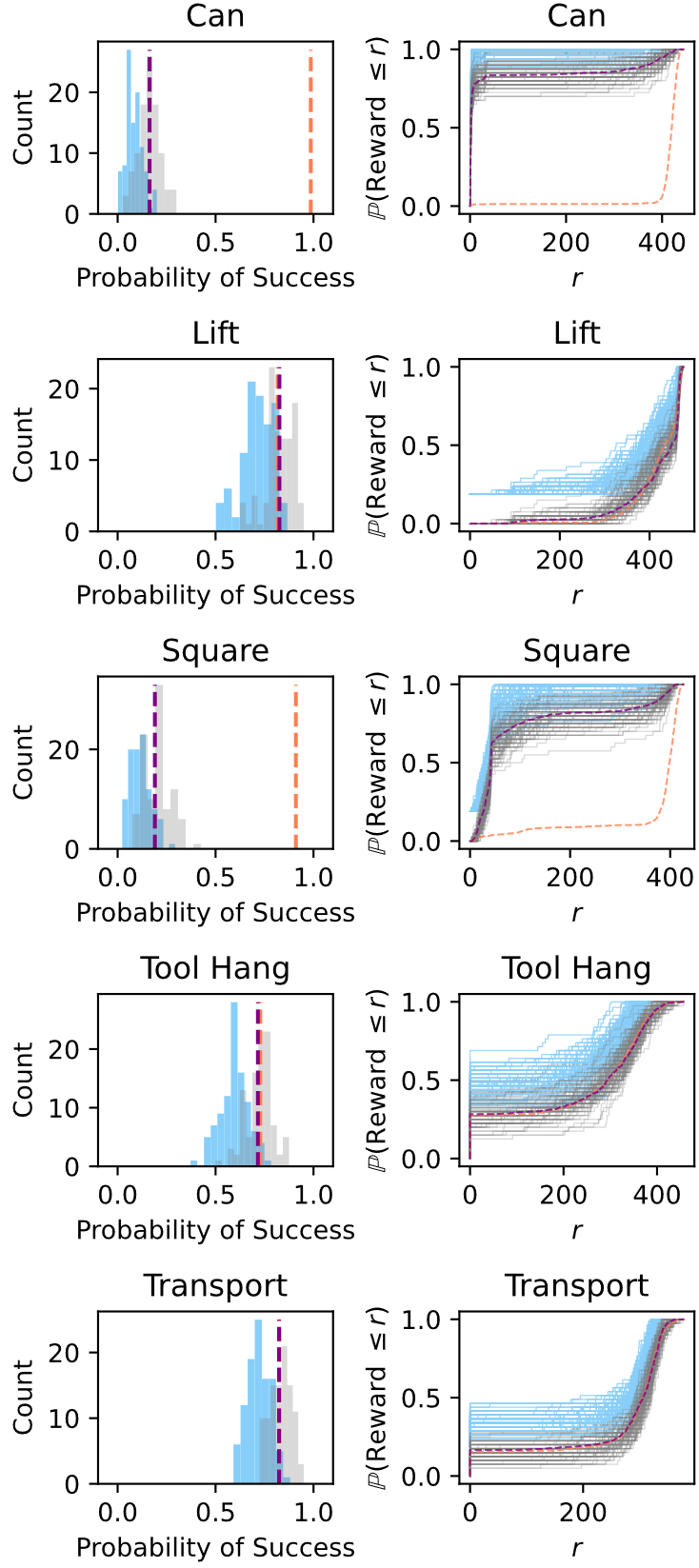}
    \caption{Visualizations of the confidence bounds for each task. Each plot shows the OOD reward distribution (purple), Monte Carlo estimates of this distribution (gray), our confidence bounds for this distribution (blue), and the ID reward distribution (orange). We use $1000$ rollouts to accurately estimate the OOD and ID reward distributions. We observe that the bounds appropriately bound the OOD reward distribution $\sim 95\%$ of the time, while a Monte Carlo estimate is overly optimistic $\sim 50\%$ of the time (since the Monte Carlo estimates are unbiased). From the ID reward distribution we can see that the distribution shifts in the Can and Square environments were harmful while the other shifts were benign.}
    \label{fig:all_tasks}
\end{figure}

In \cref{fig:all_tasks} we show our simulation results. To obtain accurate estimates of the true in-distribution (ID) (orange) and OOD (purple) performance distributions, we used 1000 policy rollouts from each environment. We use 4000 rollouts to construct confidence bounds (100 bounds generated from 40 samples each). We also show the Monte Carlo estimates of the OOD performance distribution (gray) one would obtain using 40 samples. We construct 100 bounds to empirically validate the guarantees of the bounds, but in practice one would only construct a single bound. The true ID and OOD performance distributions are typically unknown and are shown only for validating the bounds. 

From the figure we see that the distribution shifts for the Can and Square environments were quite harmful to the policy performance, while those for the other tasks were benign. Our bounds tend to accurately reflect these trends. These results reflect the unintuitive nature about when learned policies generalize, e.g., it is not clear why changing the color of the manipulated object is harmful in some cases (Can and Square) but benign in other cases (Lift and Transport). Lastly, the Monte Carlo estimates of the distributions are overly optimistic, making them unfit to be interpreted as bounds. \cref{tab:all_tasks} shows good agreement between the empirical confidence and tightness estimates with their theoretical values.

\floatsetup[table]{objectset=centering,capposition=bottom}
\begin{table}
    \begin{tabular}{lrrrrr}  
        &\multicolumn{3}{c}{\textbf{Confidence}}&\multicolumn{2}{c}{\textbf{Expected Shortage}}
        \\\cmidrule(r){2-4} \cmidrule(r){5-6}
        &Theory&Bin.&Cont.&Theory&Bin.\\\midrule
        can    &$0.95$&$0.97$&$0.95$&$0.078$&$0.079$ \\
        lift   &$0.95$&$0.94$&$0.97$&$0.115$&$0.114$ \\
        square  &$0.95$&$0.90$&$0.91$&$0.085$&$0.077$ \\
        tool hang   &$0.95$&$0.97$&$0.96$&$0.127$&$0.122$ \\
        transport   &$0.95$&$0.96$&$0.96$&$0.115$&$0.105$
    \end{tabular}
    \caption{Simulation results of the data in \cref{fig:all_tasks}. \textit{Theory} columns denote the theoretical values for confidence and expected shortage, \textit{Bin.} columns denote the empirical estimates of these quantities for the binary metric bounds, the \textit{Cont.} column denotes the empirical confidence level for the continuous metric bounds. Tightness of the continuous bounds is given by $\epsilon^*$, which is not random and thus not listed. In general, we see good agreement between the theoretical values and their empirical estimates (100 trials). With more trials we expect these estimates to converge to their theoretical values.}
    \label{tab:all_tasks}
\end{table} 

\subsection{Generalization of a Single Policy} \label{sec:hardware}
\begin{figure}
    \centering
    \includegraphics[width=\linewidth]{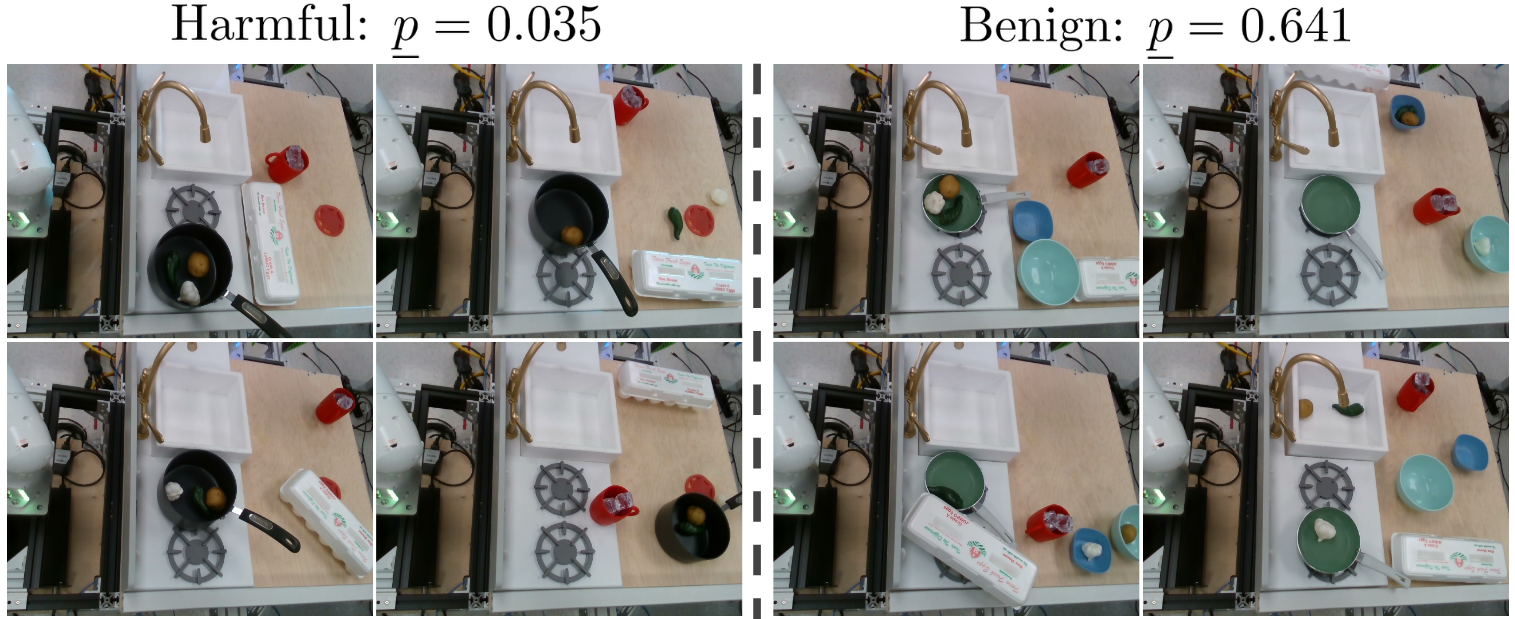}
    \caption{Snapshots of the environments used in \cref{sec:hardware}. The robot is trained to pour the ice from the red cup into the sink. Four out of $50$ initial conditions are shown for each OOD setting.}
    \label{fig:hardware_snapshots}
\end{figure}

Here we investigate the degree to which a learned policy generalizes to OOD environments. We test a diffusion policy~\cite{chi2023diffusion} trained to pour ice from a red cup into a sink. During training there are no other objects on the tabletop. We test the policy in two OOD settings shown in \cref{fig:hardware_snapshots}. The first setting is hypothesized to be a harmful distribution shift (another red object in frame) and the second is hypothesized to be a benign distribution shift (no other red objects). 

We performed $50$ policy rollouts in each setting and found $4/50$ successes in the harmful setting and $38/50$ successes in the benign setting. Applying \cref{eq:p_lo} at a $95\%$ confidence level we find $\underline{p}_{\text{harmful}} =0.035$, and $\underline{p}_{\text{benign}} = 0.641$ with $\textup{MES}=0.118$. From this, we expect $\underline{p}_{\text{harmful}}$ and $\underline{p}_{\text{benign}}$ to be reasonably close to the true success rates. The low lower-bound for the harmful setting is attributed to frequent grasp failure despite its tendency to move towards the red mug.

The bounds we find support our hypotheses, and also inform us as to the degree of performance we can assert at a $95\%$ confidence level. If a user requires a $60\%$ success rate, then we could claim with $95\%$ confidence that the benign setting is suitable for deployment whereas we could not for the harmful setting. We may then choose to retrain the policy with data from this setting and re-evaluate.

\subsection{Comparing the Generalization of Two Policies} \label{sec:comparison_ex}
In this experiment we compare two learned visuomotor manipulation policies from \cite{brohan2023rt}, where the authors test their RT-2 policy against a VC-1 policy in three settings designed to test emergent capabilities in symbol understanding, reasoning, and human recognition. We compute a lower confidence bound on the success rate of the RT-2 policy ($\underline{p}_{\text{RT-2}}$) and an upper confidence bound on the success rate of the VC-1 policy ($\overline{p}_{\text{VC-1}}$). Note, $\overline{p}_{\text{VC-1}} = 1-\underline{q}_{\text{VC-1}}$, where $\underline{q}_{\text{VC-1}}$ is a lower confidence bound on the failure rate. Each bound has confidence level $0.975$, to ensure a joint confidence of $0.95$,
\begin{gather}
    \mathbb{P}\Big[p_{\text{RT-2}} \in [\underline{p}_{\text{RT-2}}, 1]\ \cap\ p_{\text{VC-1}} \in [0, \overline{p}_{\text{VC-1}}]\Big] \ge 0.95.
\end{gather}
Then, if $\overline{p}_{\text{VC-1}} < \underline{p}_{\text{RT-2}}$ (disjoint bounds), we conclude that the RT-2 policy outperforms the VC-1 policy. Furthermore, the chances we come to this conclusion incorrectly are $\le 5\%$.

\begin{figure}[t!]
    \centering
    \includegraphics[width=0.8\linewidth]{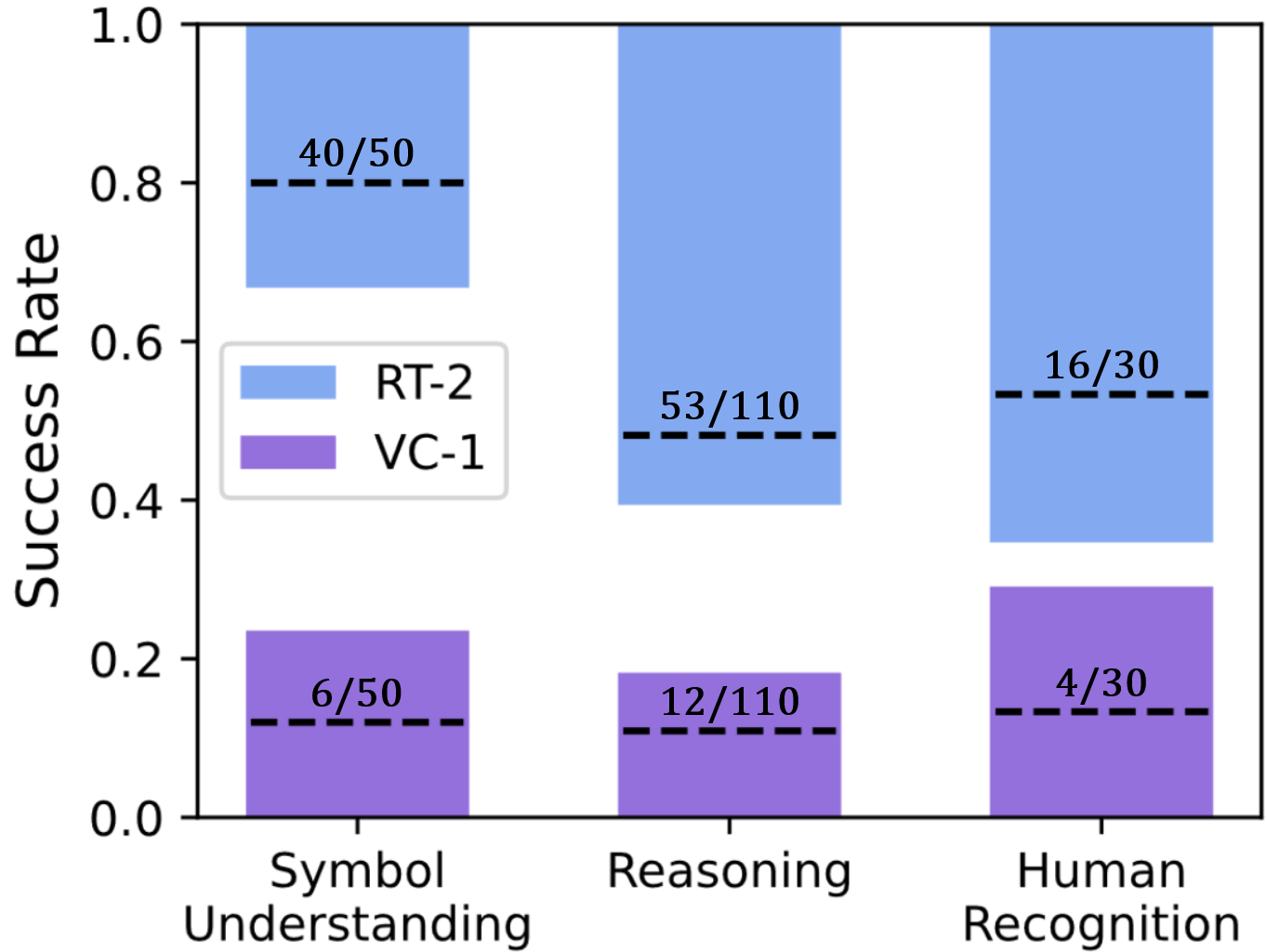}
    \caption{Confidence intervals for success rates of RT-2 (blue) and VC-1 (purple) policies. For each setting, the intervals are disjoint and we conclude, with $95\%$ confidence, that RT-2 outperforms VC-1. For each policy and setting we also show \# successes / \# trials and the corresponding empirical success rates (dashed black).}
    \label{fig:comparison}
\end{figure}

In \cref{fig:comparison}, we show the bounds for each policy in each evaluation setting. We find in each case there is enough evidence to conclude (at a $95\%$ confidence level) that the RT-2 policy outperforms the VC-1 policy. For some settings the gap between the bounds is larger, this may be due to the actual success rates being further apart, or an effect of sample size. With separate bounds for each policy, we can also conclude how well each policy performs individually, an important aspect that \textcolor{black}{standard A/B testing approaches (like Fisher's exact test) do not provide~\cite{two_sample}.} Finally, this approach easily extends to the case of continuous rewards. A lower bound on the CDF is easily computed since $D_n^+ = \sup_x \ F_n(x) - F(x)$ has the same distribution as $D_n^-$ (\cref{eq: ks_dist})~\cite{birnbaum1951}.
